\documentclass[11pt,a4paper]{article}
\usepackage[hyperref]{acl2021}
\usepackage{times}
\usepackage{latexsym}

\usepackage{times}
\usepackage{latexsym}
\usepackage{subcaption}
\usepackage{caption}
\usepackage{graphicx}
\usepackage{multirow}
\usepackage{booktabs}
\usepackage{colortbl}
\usepackage{xcolor}
\usepackage{amsmath}
\usepackage{amssymb}
\usepackage{amsfonts}
\usepackage{algorithm}
\usepackage{algorithmic}
\usepackage{float}
\usepackage{bbm}
\usepackage{kotex}
\usepackage{mathtools}
\usepackage{booktabs}
\usepackage[utf8]{inputenc}
\usepackage{enumitem}
\usepackage{amsthm}

\newcommand\modelname{LOUVRE}

\usepackage{microtype}
\let\SUP\textsuperscript

\aclfinalcopy % Uncomment this line for the final submission
\def\aclpaperid{521} %  Enter the acl Paper ID here

\newcommand\jw[1]{\textcolor{black}{{#1}}}
\newcommand\attn[1]{\textcolor{black}{{#1}}}

\title{Weakly Supervised Pre-Training for Multi-Hop Retriever}

\author{Yeon Seonwoo\SUP{$\dagger$}{\normalfont ,} Sang-Woo Lee\SUP{$\ddagger \mathsection$}, Ji-Hoon Kim\SUP{$\ddagger \mathsection$}{\normalfont ,} Jung-Woo Ha\SUP{$\ddagger \mathsection$}{\normalfont ,} Alice Oh\SUP{$\dagger$}\\
  \SUP{$\dagger$}KAIST\\
  \SUP{$\ddagger$}NAVER AI Lab, \SUP{$\mathsection$}NAVER Clova\\
  {\tt yeon.seonwoo@kaist.ac.kr}\\
  {\tt\{sang.woo.lee,genesis.kim,jungwoo.ha\}@navercorp.com}\\
  {\tt alice.oh@kaist.edu}\\
}

\date{}

\begin{document}
\maketitle
\begin{abstract}
In multi-hop QA, answering complex questions entails iterative document retrieval for finding the missing entity of the question.
The main steps of this process are sub-question detection, document retrieval for the sub-question, and generation of a new query for the final document retrieval.
However, building a dataset that contains complex questions with sub-questions and their corresponding documents requires costly human annotation.
To address the issue, we propose a new method for weakly supervised multi-hop retriever pre-training without human efforts.
\attn{Our method includes 1) a pre-training task for generating vector representations of complex questions, 2) a scalable data generation method that produces the nested structure of question and sub-question as weak supervision for pre-training, and 3) a pre-training model structure based on dense encoders.}
We conduct experiments to compare the performance of our pre-trained retriever with several state-of-the-art models on end-to-end multi-hop QA as well as document retrieval.
The experimental results show that our pre-trained retriever is effective and also robust on limited data and computational resources.
\end{abstract}

\section{Introduction}
\begin{figure*}[]
    \centering
    \includegraphics[width=0.8\linewidth]{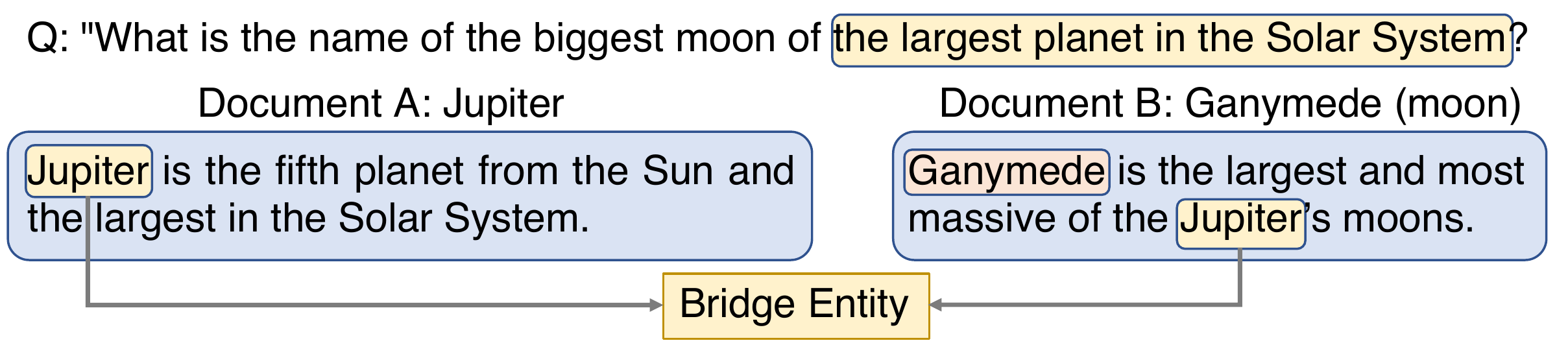}
    \vspace{-0.5em}
    \caption{An example of chain reasoning in multi-hop QA. Answering the question requires finding \textit{``the largest planet in the Solar System"} which is the bridge entity \textit{``Jupiter."} With the retrieved document A, the question has enough information to retrieve the correct answer in document B.}
    \label{fig:bridge_entity}
    \vskip -0.15in
\end{figure*}

Multi-hop QA is the task of answering complex questions that requires reasoning across multiple documents \cite{nogueira2017task, nie2019revealing, sun2019pullnet, fang-etal-2020-hierarchical, zhao2019transformer}.
The core components of multi-hop reasoning are identifying the missing entity in the question and generating a new query with the missing entity.
Figure \ref{fig:bridge_entity} shows an example of the reasoning process in multi-hop QA.
In the example, the missing entity, which we call bridge entity, of the question is \textit{``Jupiter"}.
To answer the question, the correct document for the sub-question ``the largest planet in the Solar System," should be retrieved.
Supervised training of the multi-hop QA models for these intermediate reasoning steps requires a dataset of complex questions, sub-questions, and their corresponding documents.
However, building such dataset requires costly human annotation and cannot be done at scale \cite{min2019multi, wolfson2020break}.

When there is limited annotated supervision signal, weakly supervised pre-training can be a solution \cite{devlin2019bert, liu2019roberta} which has shown effectiveness in open-domain QA \cite{lee2019latent, guu2020retrieval}.
Unlike open-domain QA, it is not trivial to apply a pre-training method to multi-hop QA due to the complexity in generating weak supervision data.
In open-domain QA, weak supervision is generated by selecting a document from a corpus and extracting a sentence from the document.
The sentence becomes a pseudo question, and the document becomes a pseudo supporting document to be predicted by retrievers.
This two-step process for weak supervision cannot be directly applied in multi-hop QA since each multi-hop question refers to multiple documents.

\attn{
In this paper, we propose a novel weakly supervised pre-training method for multi-hop retriever, \textbf{\modelname{}} (\textbf{L}earning fr\textbf{O}m m\textbf{U}lti-hop \textbf{V}ariation of document \textbf{RE}lations).
}
Our method contains three core elements: 1) a pre-training task, 2) a scalable method to generate pre-training data with weak supervision, and 3) a model to pre-train a retriever for multi-hop QA.
Specifically, we define a task for pre-training, ``Next Document Prediction" (NDP), which is to retrieve documents for sub-questions.
We then propose ``Bridge Entity Re-Phrasing" to generate the pre-training data.
``Bridge Entity Re-Phrasing" generates complex questions that contain sub-questions of the bridge entities and their corresponding documents.
To generate a complex question using a bridge entity without human effort, we use two documents connected by Wikipedia hyperlinks.
The hyperlinked entity becomes the bridge entity, and the introductory phrase of the entity becomes the sub-question in the complex question.
This approach enables our weak supervision data generation to be scalable, as shown in Figure 2.
We use a dense retriever consisting of a question encoder and a document encoder for the pre-trained model structure \cite{karpukhin-etal-2020-dense, xiong2020answering}.
The two encoders calculate vector representations of questions and documents.
Document retrieval is performed by comparing the vectors with MIPS (maximum inner product search).

Pre-training multi-hop retriever with our weak supervision method brings three \jw{benefits}: significant performance improvement, robustness on few-shot settings, and computational efficiency.
\jw{We evaluate our weakly supervised pre-trained retriever with two types of experiments on \normalsize{H}\small{OTPOT}\normalsize{QA} dataset: supporting documents prediction and end-to-end multi-hop QA.}
In both experiments, \modelname{} outperforms previous multi-hop retrievers.
Also, we fine-tune \modelname{} on 1\% of training data and show that the performance of \modelname{} is comparable to the baselines.
We evaluate the performance of \modelname{} according to the computational efficiency.
The results show that \modelname{} requires less inference time than baselines.

Contributions of this paper are as follows: 1) we propose a novel scalable weakly supervised pre-training method for multi-hop retrievers, 2) we provide the implementation of \modelname{} and the pre-trained checkpoint publicly available \footnote{\attn{\url{https://github.com/yeonsw/LOUVRE}}}, 3) we show the effect of our pre-training method in multi-hop QA with various experimental results.

\section{Related Work}
\paragraph{Distant Supervision in Open-Domain QA:}
Many open-domain QA datasets only provide question-answer pairs; some also provide weakly annotated supporting documents, but they are predicted by simple heuristics \cite{joshi2017triviaqa, berant2013semantic}.
\jw{Document retrieval has suffered from insufficient strong supervision issues.}
Hence, document retrieval has suffered from lack of strong supervision.
To resolve this issue,
\attn{\newcite{karpukhin-etal-2020-dense} use a document retrieved by TF-IDF as the supporting document of the given question.}
\attn{Weak supervision is also an effective method in the distant supervision setting of open-domain QA.}
\newcite{lee2019latent} use ICT (inverse cloze task) to generate pseudo question-document pairs and pre-train their retriever.
They select documents from Wikipedia and extract sentences from the documents.
The selected sentence-document pairs become pseudo-question-document pairs.
\newcite{guu2020retrieval} propose a pre-training method for a language model that uses knowledge retriever (document retriever).
They train the knowledge retriever only with the language modeling loss without using any supervision signal of supporting documents.
Although pre-training methods show effectiveness in open-domain QA, they are limited to single-hop retrievers.

\begin{figure*}[]
    \centering
    \includegraphics[width=0.99\linewidth]{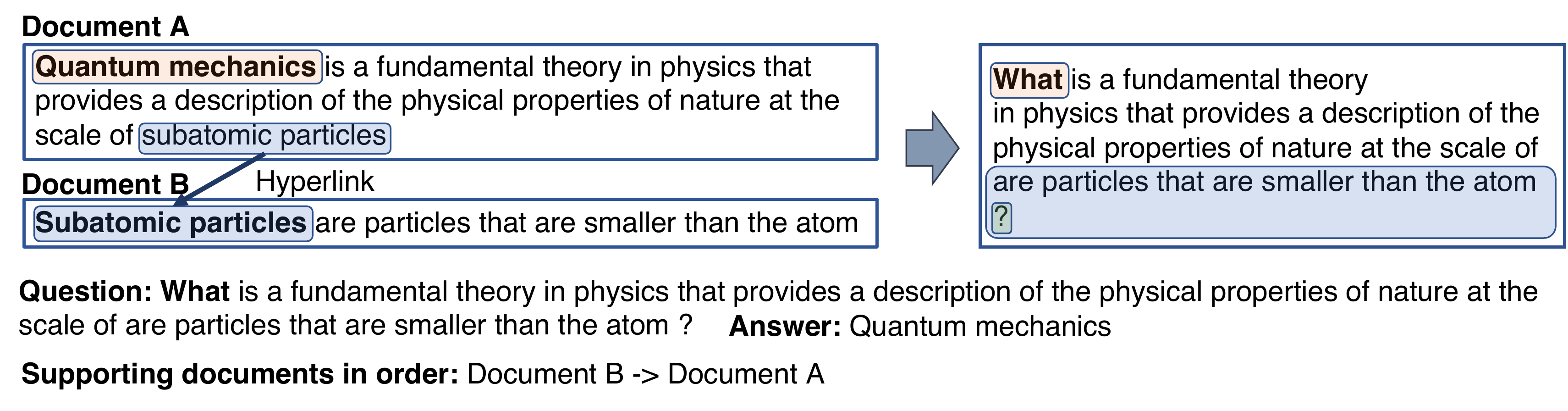}
    \caption{Proposed pre-training data generation process. Two documents connected by Wikipedia hyperlink are selected. In ``Bridge Entity Re-Phrasing" process, document B which describes the entity ``subatomic particles" is used to re-phrase the entity in document A. After replacing the answer entity, ``Quantum mechanics", the complex question and its corresponding supporting documents are generated.}
    \label{fig:weak_supervision}
    \vspace{-1em}
\end{figure*}

\paragraph{Multi-Hop QA:}
To overcome the lack of supervision signal in multi-hop QA, weak supervision methods have been proposed.
\newcite{qi2019answering} propose a sub-question generation method.
They use heuristically generated pseudo-questions as supervision for the question generation model.
\newcite{perez-etal-2020-unsupervised} generate weak supervision for question decomposition by mapping a complex question to multiple single-hop questions in existing QA datasets.
\attn{They use complex questions in \normalsize{H}\small{OTPOT}\normalsize{QA} \cite{yang2018hotpotqa} and single-hop questions in SQuAD 2.0 \cite{rajpurkar2018know}.}
Another method to train multi-hop QA models without human annotated datasets is by taking two simple questions and generating a complex question 
\cite{pan2020unsupervised}.
\attn{They generate complex questions with GPT-2 \cite{radford2019language} fine-tuned on SQuAD1.1.}
Our work improves upon previous research by providing a more general method that leverages a large open corpus with retriever pre-training.

\section{Method}
We propose an effective and scalable pre-training method that provides weak supervision of the complex questions with sub-questions and their corresponding supporting documents.

\subsection{Next Document Prediction}
We propose the ``Next Document Prediction" (NDP) task for pre-training.
NDP refers to the process of recurrent document retrieval used in \cite{qi2019answering, asai2019learning, xiong2020answering}.
We apply the common definitions in the existing studies to our ``Next Document Prediction" task.
We define NDP as the task that predicts documents in the reasoning sequence $[d_1, ..., d_n]$ recurrently as follows:
\begin{equation}
    d_k = \text{Retriever}(q, D_{k-1}),
\end{equation}
where $q$ is a question, $d_k$ is a predicted document at step $k$, and $D_{k-1}$ is a set of documents retrieved in the previous steps, $\{d_1, ..., d_{k-1}\}$.

\subsection{Bridge Entity Re-Phrasing}\label{sec:ber}
Our pre-training requires a dataset of questions, sub-questions, and their corresponding reasoning chains (i.e., a sequence of documents).
We propose ``Bridge Entity Re-Phrasing" for generating this pre-training dataset.
``Bridge Entity Re-Phrasing" takes two steps: entity selection and re-phrasing.
Figure \ref{fig:weak_supervision} provides an overview of our data generation process.
We provide the detailed description of ``Bridge Entity Re-Phrasing" in the following paragraphs.

The ``Bridge Entity Re-Phrasing" process requires informative entities and the description of the entity.
\attn{We assume that an entity with a Wikipedia hyperlink is an informative entity.}
\attn{Also, hyperlink entities often have Wikipedia articles describing the entities.}
\attn{The hyperlink entity becomes the bridge entity.}
In Figure \ref{fig:weak_supervision}, document A and document B are connected with the bridge entity, \textit{``subatomic particles"}.
We re-phrase the selected entity with the first line of the document.
In Figure \ref{fig:weak_supervision}, \textit{``subatomic particles"} is re-phrased with the first line in document B.
\attn{When the bridge entity appears in the question, multi-hop retrievers easily find the bridge document using only the word.}
\attn{To prevent this issue, we remove the bridge entity from document B.}
\attn{The generated document becomes the document to be used for question-answer pair generation.}

Generating a question-answer pair from a single document has been studied by pre-training research in open-domain QA \cite{lee2019latent, guu2020retrieval}.
We extend their work to generate questions, reasoning chains, and answers.
We \jw{randomly} select \jw{an entity} from the merged document and replace the entity with the word \textit{``what"}.
In Figure \ref{fig:weak_supervision}, ``Quantum mechanics" is the entity word and replaced with ``what."
The new sentence becomes a pseudo-question, and the replaced entity becomes the answer.
Since document B contains the bridge entity, the pseudo-question reasoning chain becomes $[\text{document B}, \text{document A}]$.

\subsection{Model Architecture}
Model structure for our pre-training is subject to two requirements: general model structure and recurrent retrieval.
We use multi-hop dense retriever \cite{xiong2020answering} which meets the two requirements and is based on the DPR (dense passage retriever) \cite{karpukhin-etal-2020-dense}.
DPR consists of a question encoder $E_Q$ and a document encoder $E_D$, both of which are based on RoBERTa-base.
Documents are retrieved by MIPS (maximum inner product search) with similarity between the question vectors and the document vectors as follows:
\begin{equation} \label{eq:mips}
    \text{sim}(q, d) = E_Q(q)^\intercal E_D(d).
\end{equation}
MDR retrieves documents recurrently by taking the previously retrieved documents as input.
MDR concatenates the question $q$ and the retrieved documents $\{d_1, ..., d_{k-1}\}$ and calculates a question vector for $k$-th step as follows: 
\begin{equation}\label{eq:recurrent}
    d_k = \text{argmax}_{d}(E_Q(q, d_1, ..., d_{k-1})^\intercal E_D(d)),
\end{equation}
where $d$ is a document in the corpus.

We train the dense encoder to assign the highest probability for the ground truth document among the documents in the huge corpus.
The loss function for our pre-training is as follows:
\begin{equation}\label{eq:loss}
    \begin{split}
        L&_{\text{NBP}}(q_k, d_k) = \\
        &-\log \frac{e^{\text{sim}(q_k, d_k)}}{e^{\text{sim}(q_k, d_k)} + \sum_{d \in \text{neg}(d_k)} e^{\text{sim}(q_k, d)}},
    \end{split}
\end{equation}
where $q_k$ is a concatenation of $q$ and $D_{k-1}$, and $\text{neg}(d_k)$ is a set of documents excluding $d_k$.
Since computing the softmax over the whole corpus is computationally expensive, we use in-batch negatives for $\text{neg}(d_k)$ \cite{karpukhin-etal-2020-dense}.

\section{Experimental Setup}\label{sec:exp_set}
\subsection{Pre-Training Details}
We generate our pre-training data from 5,233,329 Wikipedia articles provided by \newcite{yang2018hotpotqa}.
We select all sentences that contain at least one hyperlinked entity to generate pseudo questions and randomly select ``answer" entities from the sentences \footnote{We use spaCy for entity recognition}.
Our data generation process builds 13.9 million question-document-answer triples.
We pre-train our dense retriever with a batch size of 256 for 200K+ steps.
We use Adam with a warm-up ratio of $0.1$ and set the learning rate to $2\times 10^{-5}$.
We use a machine with eight V100 (32G) GPUs.

\subsection{Fine-Tuning Details}
We use TF-IDF negatives in addition to in-batch negatives for fine-tuning as in \newcite{karpukhin-etal-2020-dense, xiong2020answering}.
We set the number of TF-IDF negatives to 2.
We use the Adam optimizer with a warm-up rate of $0.1$ and set the learning rate to $2\times10^{-5}$.
We set the batch size to 32, the number of epochs to 15.
\attn{To achieve better performance, we additionally fine-tune our model with another hyper-parameter setting: a batch size of 150 and a longer training time of 50 epochs.}

\subsection{Tasks}
\noindent\emph{\textbf{Supporting Document Prediction:}}
In this task, retrievers and rerankers predict supporting documents for each question in \normalsize{H}\small{OTPOT}\normalsize{QA} dataset \cite{yang2018hotpotqa}.
\attn{The models predict possible combinations of supporting documents.
Formally, when a question, $q_i$, has been taken as input of the model, the models yield a ranked list of K-sets of documents, $L_i = [\{d_a, d_b\}]_{n=1}^K$.
Each $\{d_a, d_b\}$ is a pair of candidate supporting documents.
}
In \normalsize{H}\small{OTPOT}\normalsize{QA}, the number of supporting documents is fixed to 2.
We use the 5 million Wikipedia articles as the knowledge source.

\noindent\emph{\textbf{End-to-End Multi-Hop QA:}}
We evaluate the supporting facts prediction performance and the answer prediction performance of \modelname{} on \normalsize{H}\small{OTPOT}\normalsize{QA} full wiki setting \cite{yang2018hotpotqa}.

\begin{table}[]
\centering
\begin{tabular}{@{}lcccc@{}}
\toprule
                 & TF-IDF     & Wiki       & RR         & Eff        \\ \midrule
\modelname{}     &            &            &            &            \\
- eff            & \checkmark & \checkmark &            & \checkmark \\
- Wiki           &            & \checkmark &            &            \\
- reranking      &            &            & \checkmark &            \\
- reranking-Wiki &            & \checkmark & \checkmark &            \\ \bottomrule
\end{tabular}
\caption{
The five variations of \modelname{}. Each column represents sparse retrieval (TF-IDF), Wikipedia hyperlinks (Wiki), reranking (RR), and efficient fine-tuning (Eff).
``Eff" represents whether the model uses efficient hyper-parameter setting, a batch size of 32 and a number of epochs of 15.
}
\label{tab:model_types}
\end{table}

\subsection{Multi-Hop Retrieval Strategy}\label{sec:retrieving_strategy}
\attn{We propose five variations of \modelname{} based on existing multi-hop retrieval strategies.
Multi-hop document retrievers leverage three strategies for performance improvement and computational efficiency: sparse retrieval methods such as TF-IDF \cite{nie2019revealing}, Wikipedia hyperlinks \cite{asai2019learning}, and reranking \cite{xiong2020answering}.
Sparse retrieval methods select a small number of candidate documents relevant to the given question and are used to narrow down the search space of dense retrievers.
We use TF-IDF and keyword matching as \newcite{nie2019revealing} to retrieve 200 candidate documents.
Existing multi-hop retrievers select reasoning paths (document chains) from documents connected with Wikipedia hyperlinks.
We iteratively select the next-hop documents from the documents connected with the previously retrieved documents.
Rerankers take the candidate reasoning paths (pairs of documents) from the retriever and predict the most probable reasoning path.
We use the reranker proposed by \newcite{xiong2020answering}.
Table \ref{tab:model_types} shows the detailed information of these five variations.
}

\subsection{Baselines}\label{sec:baselines}
\attn{We compare our model to two types of multi-hop retrievers with and without reranking.
For retrievers, we use TF-IDF, DPR \cite{karpukhin-etal-2020-dense}, Cognitive Graph \cite{ding2019cognitive}, \normalsize{G}\small{OLD}\normalsize{E}\small{N} \normalsize{Retriever} \cite{qi2019answering}, and MDR \cite{xiong2020answering}.
For rerankers, we use SemanticRetrievalMRS \cite{nie2019revealing}, PathRetriever \cite{asai2019learning}, MDR-reranking \cite{xiong2020answering}, and HopRetriever \cite{li2020hopretriever}.
We describe detailed experimental settings of each baseline in Appendix \ref{sec:appendix_baselines}.
}

\subsection{Metric}
\attn{We use five evaluation metrics: EM, F1, R@K, PathR@K, and AR@K.}
EM and F1 measure answer prediction and supporting fact prediction performance of multi-hop QA models ~\cite{yang2018hotpotqa}.
\attn{In addition to R@K, which measures the performance of supporting document prediction, we use another metric PathR@K to evaluate how well the retriever predicts the entire set of supporting documents.}
Since the readers predict answers by reading each path, PathR@K is a more appropriate estimate of answer prediction.
The definitions of R@K and PathR@K are:
\begin{equation}
    \begin{split}
        \text{R@K} &= \mathbbm{1}(G \subseteq D)\\
        \text{PathR@K} &= \mathbbm{1}((G = P_1) \lor ... \lor (G = P_K)),\\
    \end{split}
\end{equation}
where $G=\{g_i, g_j\}$ is the set of ground truth supporting documents, $D$ is the set of retrieved documents, and $P_i = \{d_a, d_b\}$ is \attn{a reasoning path ranked at $i$.}

In our experimental setting, $D$ is set to all documents in $\bigcup_{i=1}^{K/2} P_i$.
\attn{
AR@K measures the percentage of predictions that at least one passage in the top $K$ predicted paths contains the answer text.
}
\begin{table*}[]
\centering
\begin{tabular}{@{}llccccc@{}}
\toprule
                           & \multicolumn{1}{c}{}         & \multicolumn{1}{l}{Wiki link} & PathR@1       & R@10          & R@20          & AR@1          \\ \midrule
\multirow{9}{*}{Retriever} & TF-IDF                       &                               & 9.8           & 27.6          & 35.1          & 43.4          \\
                           & DPR                          &                               & 25.2          & 45.4          & 52.1          & -             \\
                           & Cognitive Graph              &                               & 57.8          & -             & -             & 76.0          \\
                           & GoldEnRetriever              &                               & -             & 75.4          & -             & -             \\
                           & MDR                          &                               & 65.9          & 77.5          & 80.2          & 75.4          \\ \cmidrule(l){2-7} 
                           & \modelname{}-eff (1\%)             & \checkmark                         & 53.5          & 75.5          & 80.0          & 72.3          \\
                           & \modelname{}-eff                   & \checkmark                         & 65.3          & 80.4          & \textbf{83.0} & \textbf{78.9} \\
                           & \modelname{}                       &                               & 67.0          & 77.8          & 80.3          & 76.3          \\
                           & \modelname{}-Wiki                & \checkmark                         & \textbf{69.5} & \textbf{80.7} & 82.4          & 77.8          \\ \midrule
\multirow{7}{*}{Reranker}  & SemanticRetrievalMRS         & \checkmark                         & 63.9          & 81.7          & 82.1          & 77.9          \\
                           & PathRetriever                & \checkmark                         & 75.7          & 82.4          & -             & 87.5          \\
                           & MDR-rerank                &                               & 81.2          & 86.4          & 86.6          & 88.2          \\
                           & HopRetriever (w/o Wiki link) &                               & 66.2          & 78.8          & -             & 76.3          \\
                           & HopRetriever                 & \checkmark                         & 82.5          & 88.6          & -             & 86.8          \\ \cmidrule(l){2-7} 
                           & \modelname{}-rerank           &                               & 83.2          & 89.1          & 89.7          & \textbf{90.0} \\
                           & \modelname{}-rerank-Wiki    & \checkmark                         & \textbf{83.5} & \textbf{89.5} & \textbf{90.1} & 89.8          \\ \bottomrule
\end{tabular}
\caption{
    \attn{
        Performance of retrievers. 
        \modelname{} outperforms baseline retrievers and rerankers.
        Wiki link denotes whether the model uses Wikipedia hyperlinks.
    }
}
\label{tab:retriever}
\end{table*}

\section{Results \& Discussion}
\attn{\modelname{} overcomes limited supervision in multi-hop QA with our weakly supervision data.
Training with additional data brings progress in three ways: overall retrieval performance improvement, 2) robustness on a few-shot setting, and 3) overall improvement in the end-to-end multi-hop QA.
We verify these improvements on the two tasks described in section \ref{sec:exp_set}: supporting document prediction and end-to-end multi-hop QA.
}

\subsection{Supporting Document Prediction}\label{sec:sup}
\attn{In this experiment, we demonstrate the efficacy of \modelname{} with document retrieval experiments.
First, we show the performance gain that comes from using our pre-trained model.
Then, we show that the result becomes more significant in few-shot settings.}

\noindent\emph{\attn{\textbf{Effect of Our Pre-Training:}}}
\attn{We compare \modelname{} with MDR which is a multi-hop retriever fine-tuned on RoBERTa-base.
We use the same fine-tuning method as MDR but initialize the parameters with \modelname{}.
Table \ref{tab:retriever} shows the results.
\modelname{} achieves 1.1\% absolute performance improvement than when using RoBERTa (65.9); PathR@1 of \modelname{} is 67.0.
Also, \modelname{} outperforms MDR in other evaluation metrics.
In reranking experiments, we use the same reranking model as MDR-rerank.
The only difference between \modelname{}-rerank and MDR-rerank is the parameter initialization method in the fine-tuning step same as the retriever experiment.
These results show that our pre-training method is effective even after reranking; PathR@1 of \modelname{}-rerank is 83.2, and PathR@1 of MDR-rerank is 81.2.
}

\noindent\emph{\textbf{Weak Supervision and Training Time:}}
\modelname{}'s pre-training method uses additional training with the multi-hop weak supervision dataset and results in the performance improvement shown above.
\attn{To verify that the performance gap between RoBERTa and \modelname{} is not from the additional training time that \modelname{} uses in pre-training, we train RoBERTa with much a longer training time, 50 epochs, and compare with \modelname{}.
}

\begin{figure}[]
    \centering
    \includegraphics[width=0.99\linewidth]{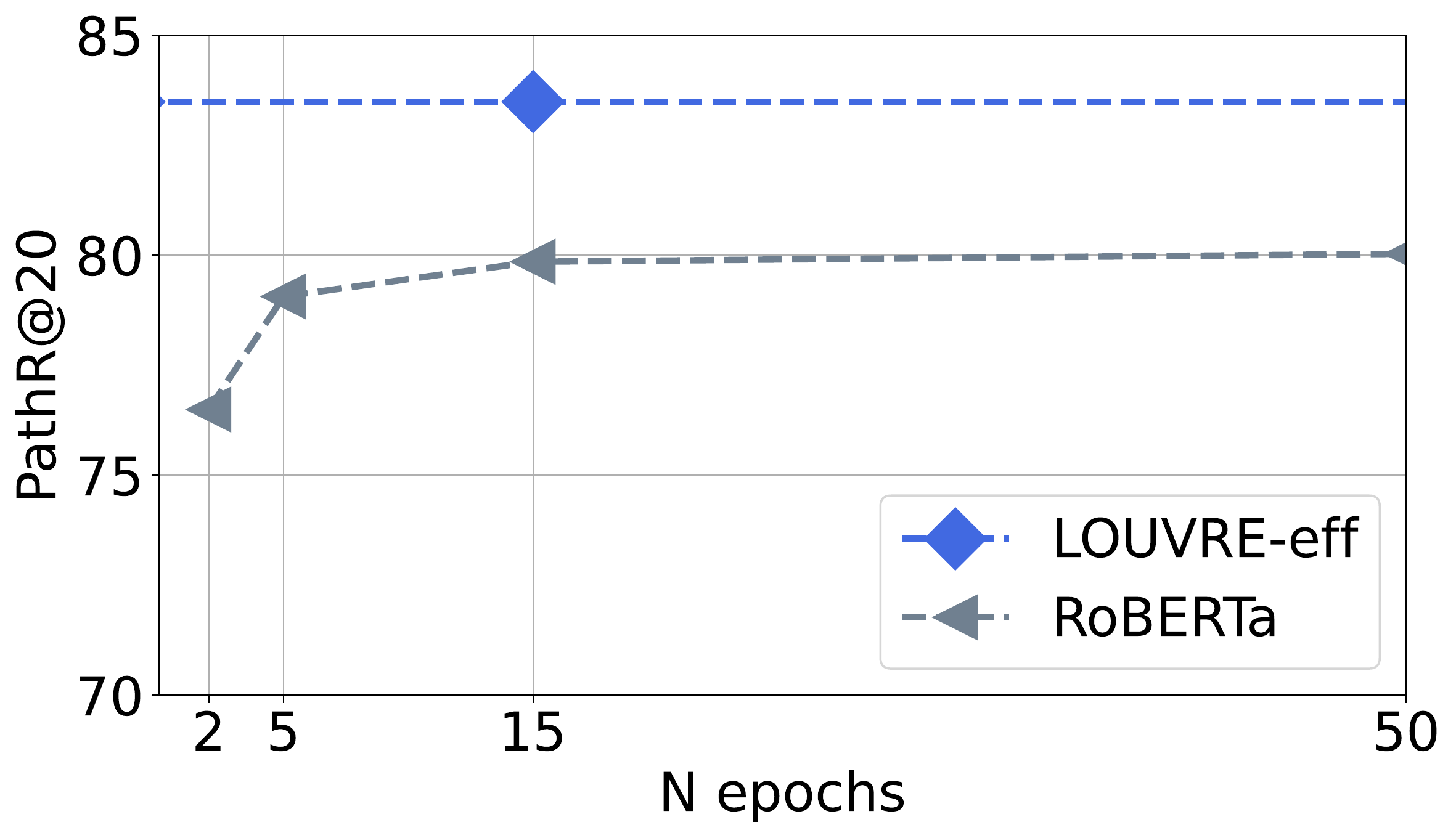}
    \caption{Comparison of \modelname{}-eff fine-tuned for 15 epochs with RoBERTa fine-tuned for 2 to 50 epochs. The performance of RoBERTa flattens at 15 epochs.}
    \label{fig:ft_n_epoch}
    \vspace{-1em}
\end{figure}

In Figure \ref{fig:ft_n_epoch}, we show the performance of RoBERTa fine-tuned for $\{2, 5, 15, 50\}$ epochs and the performance of \modelname{} fine-tuned for 15 epochs.
The performance of RoBERTa stabilizes at approximately 80\% in terms of PathR@20 after 15 epochs.
This result shows that the main factor of the performance improvement from our pre-training method is not merely from longer training time but the unique information for multi-hop retrievers provided by our weak supervision.

\begin{table}[]
\centering
\begin{tabular}{@{}lcccc@{}}
\toprule
\multicolumn{1}{c}{} & \multicolumn{2}{c}{R}         & \multicolumn{2}{c}{PathR}     \\
\multicolumn{1}{c}{} & @10           & @20           & @8            & @20           \\ \midrule
\modelname{}-eff     & \textbf{80.4} & \textbf{83.0} & \textbf{81.9} & \textbf{83.5} \\
- 1\% train data     & 75.5          & 80.0          & 77.1          & 81.5          \\
- zeroshot           & 44.8          & 55.8          & 48.3          & 59.0          \\
- w/o pre-training   & 75.7          & 78.5          & 76.9          & 79.8          \\ \midrule
DPR-zeroshot         & 39.0          & 51.6          & 42.8          & 56.6          \\ \bottomrule
\end{tabular}
\caption{Supporting document prediction performance of retrievers.}
\label{tab:retriever_eff}
\end{table}

\noindent\emph{\attn{\textbf{Retrieval Performance of Variations of \modelname{}-eff:}}}
\attn{Table \ref{tab:retriever} shows the effect of using \modelname{}.
Similar results are observed in \modelname{}-eff.
Table \ref{tab:retriever_eff} shows the same experiments as Table \ref{tab:retriever} but in efficient fine-tuning setting, a small batch size and a short train time.
We compare \modelname{}-eff with the retrieval performance of \modelname{}-eff without our pre-training method, which is fine-tuned on RoBERTa.
Applying our pre-training method increases the retrieval performance by 4.7\% point (R@10); R@10 of \modelname{}-eff is 80.4 and R@10 of \modelname{}-eff without pre-training is 75.7.
Taking the results in Table \ref{tab:retriever} (the performance gain from our method with a big batch size/train epochs is 1.1) and the results in Table \ref{tab:retriever_eff}, we see the performance gain increases as there is more limitation on computation time.
}

\begin{figure}[t]
    \centering
    \begin{subfigure}[b]{0.47\textwidth}
        \centering
        \includegraphics[width=0.99\linewidth]{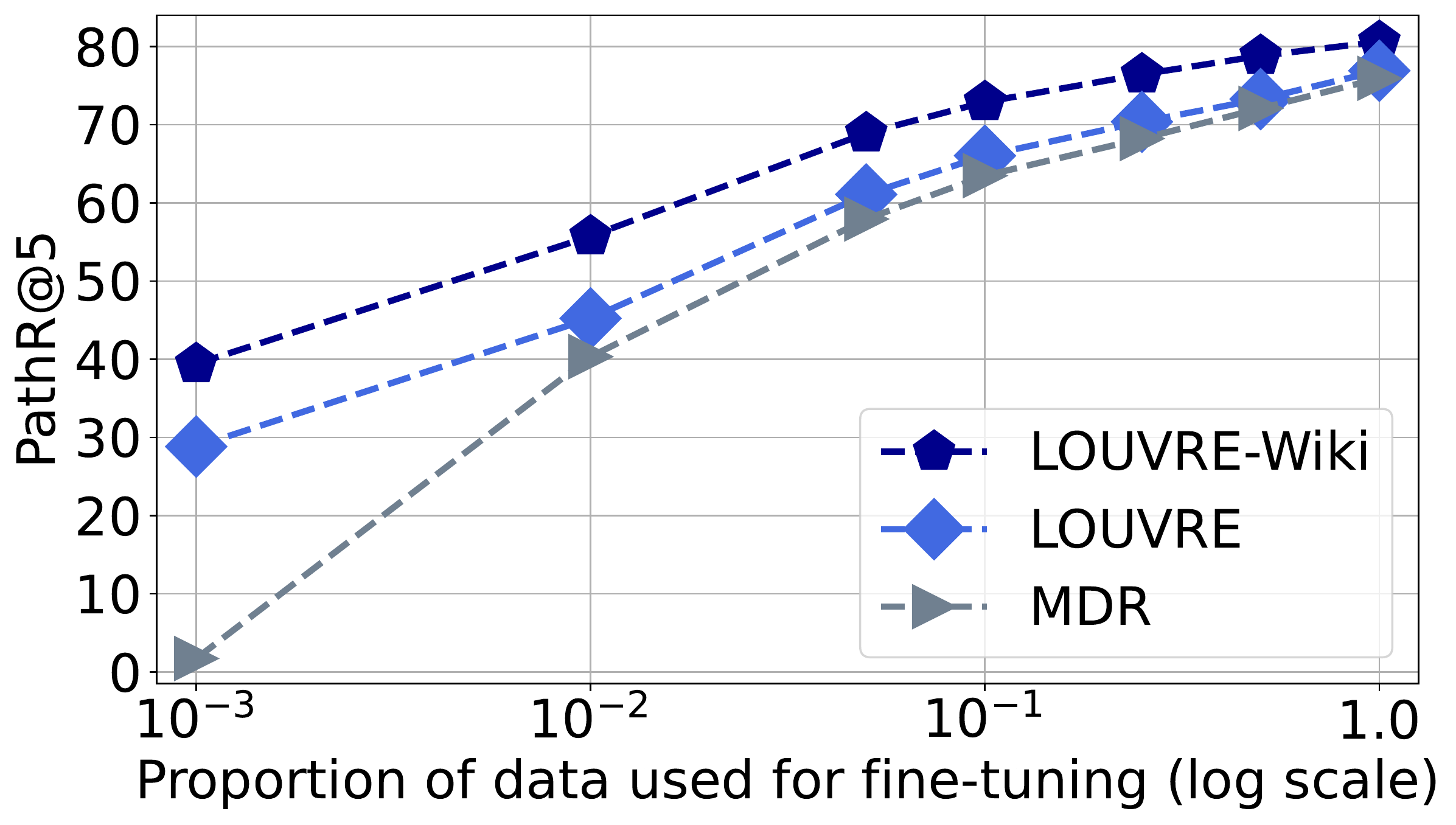}
    \end{subfigure}
    \begin{subfigure}[b]{0.47\textwidth}
        \centering
        \includegraphics[width=0.99\linewidth]{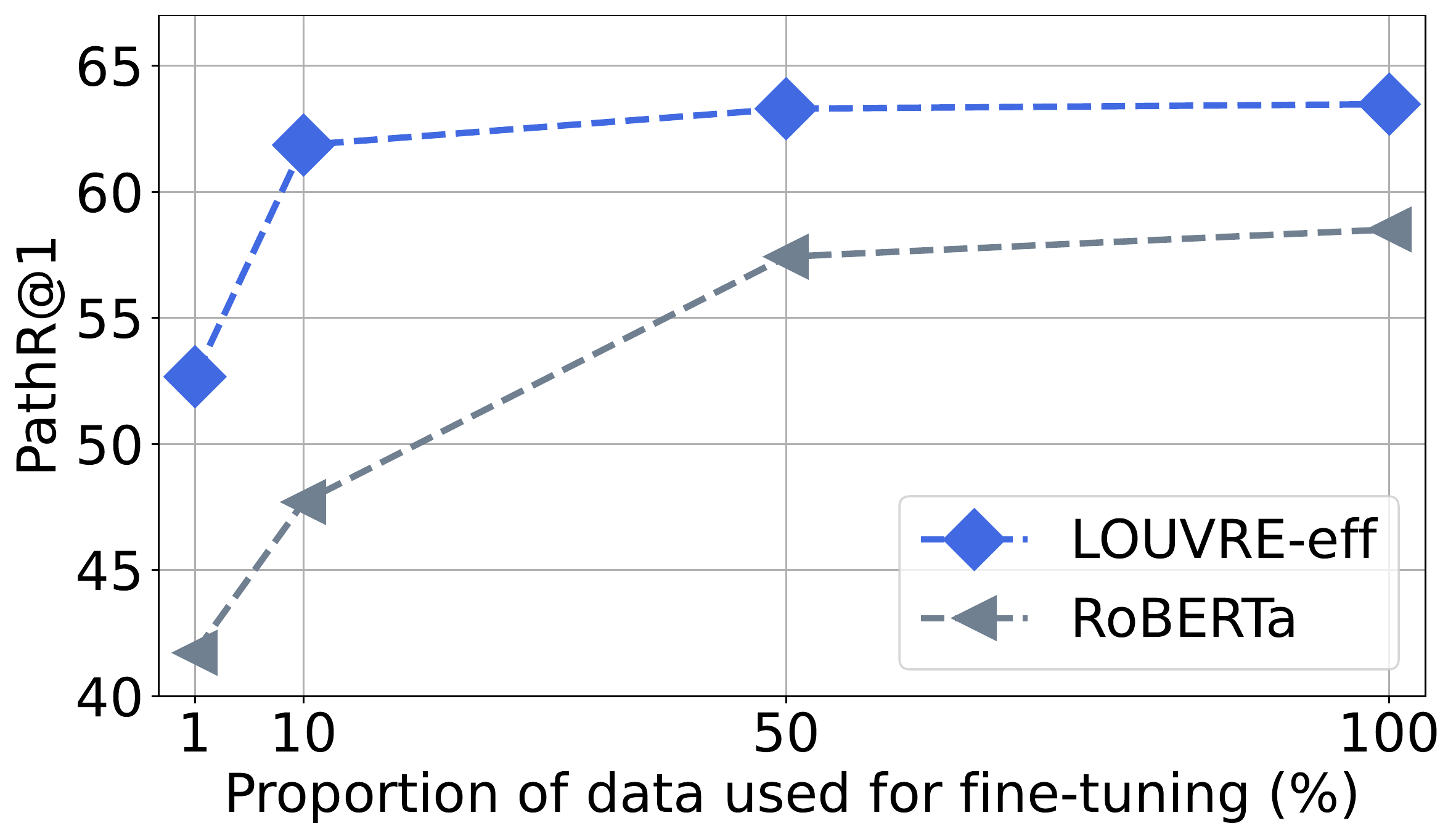}
    \end{subfigure}
    \caption{Retrieval performance of \modelname{}, \modelname{}-eff, and RoBERTa. The only difference between each model in the comparison models, (\modelname{} and MDR) and (\modelname{}-eff and RoBERTa), is the pre-trained model being used for parameter initialization.}
    \label{fig:data_eff}
\end{figure}

\noindent\emph{\attn{\textbf{Robustness on Few-shot Settings:}}}
\attn{Pre-training alleviates the model's drastic performance drop when the number of training data is insufficient.
We demonstrate the robustness of \modelname{} on few-shot settings with different sizes of train data.
We fine-tune \modelname{} and MDR on a small portion of the \normalsize{H}\small{OTPOT}\normalsize{QA} train set within 0.1\% to 100\%.
Figure \ref{fig:data_eff} shows that the performance gap between \modelname{} and MDR increases as the size of train data decreases.
When we use 0.1\% of \normalsize{H}\small{OTPOT}\normalsize{QA} train data, almost 30\% of \modelname{}'s predictions contain correct supporting documents; the performance of MDR with the same amount of train data is close to 0.
We conduct the same experiment with \modelname{}-Wiki and verify that using Wikipedia hyperlinks improves the robustness on a few-shot setting by 10.5\% point in terms of PathR@5 when there is only 0.1\% train data.
}

\attn{We conduct the same experiment with a small batch size of 32 and 10 epochs.
Figure \ref{fig:data_eff} illustrates the retrieval performance of \modelname{}-eff and \modelname{}-eff without our pre-training (RoBERTa) depending on the proportion of the data used for fine-tuning.
It is worth noting that \modelname{}-eff fine-tuned with 10\% data outperforms RoBERTa with 100\% and shows little performance degradation compared to fine-tuning with 100\%.
We report the detailed results of \modelname{}-eff (1\%) in Table \ref{tab:retriever}.
Table \ref{tab:retriever} shows that \modelname{}-eff (1\%) achieves comparable performance to MDR trained on full data with a larger batch size and a longer train time; R@10 of \modelname{}-eff (1\%) is 2.0\% point lower than R@10 of MDR, which is 97\% of the performance of MDR.
}

\begin{figure}[t]
    \centering
    \begin{subfigure}[b]{0.49\textwidth}
        \includegraphics[width=0.99\linewidth]{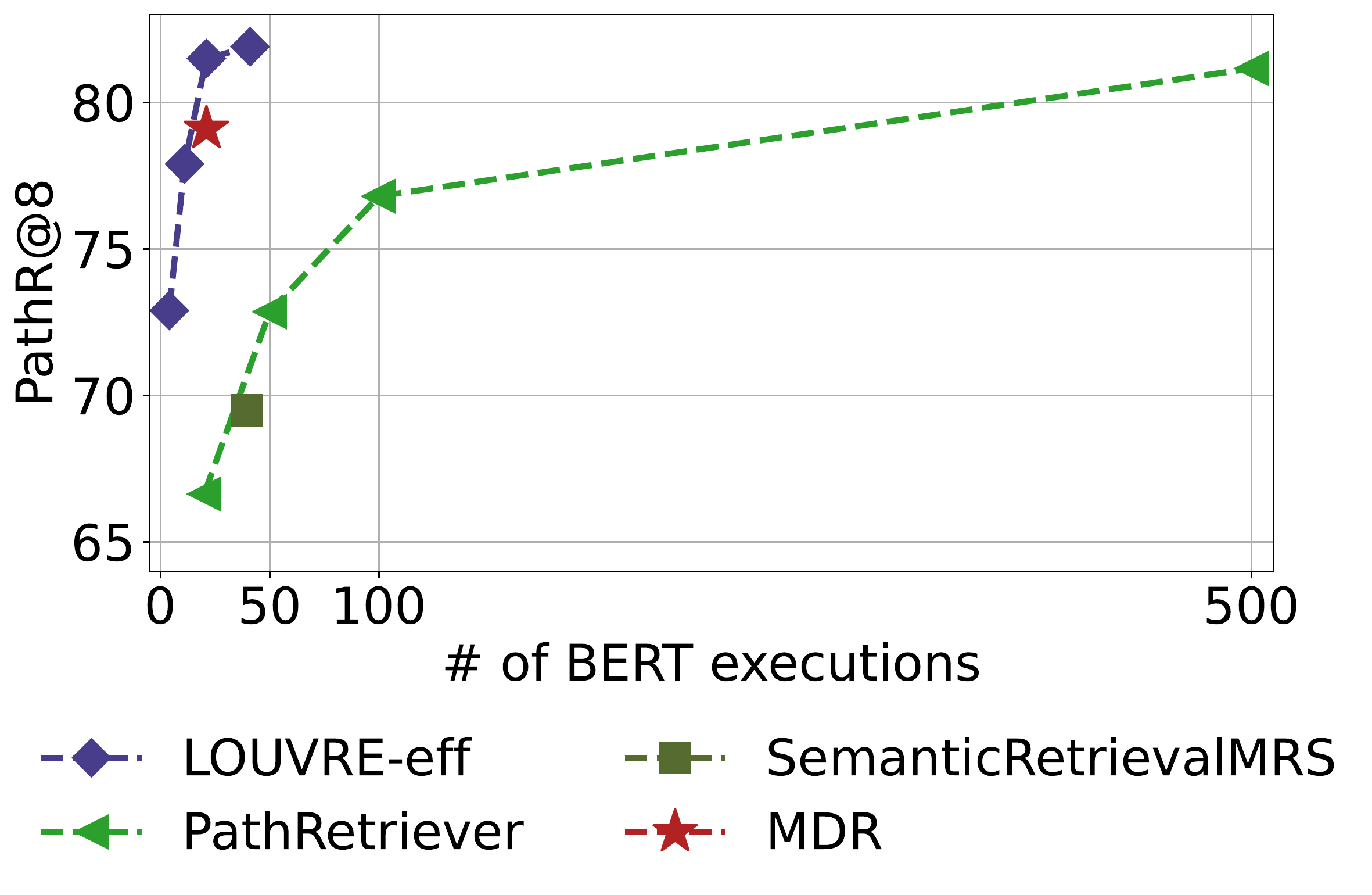}
    \end{subfigure}
    \begin{subfigure}[b]{0.47\textwidth}
        \includegraphics[width=0.99\linewidth]{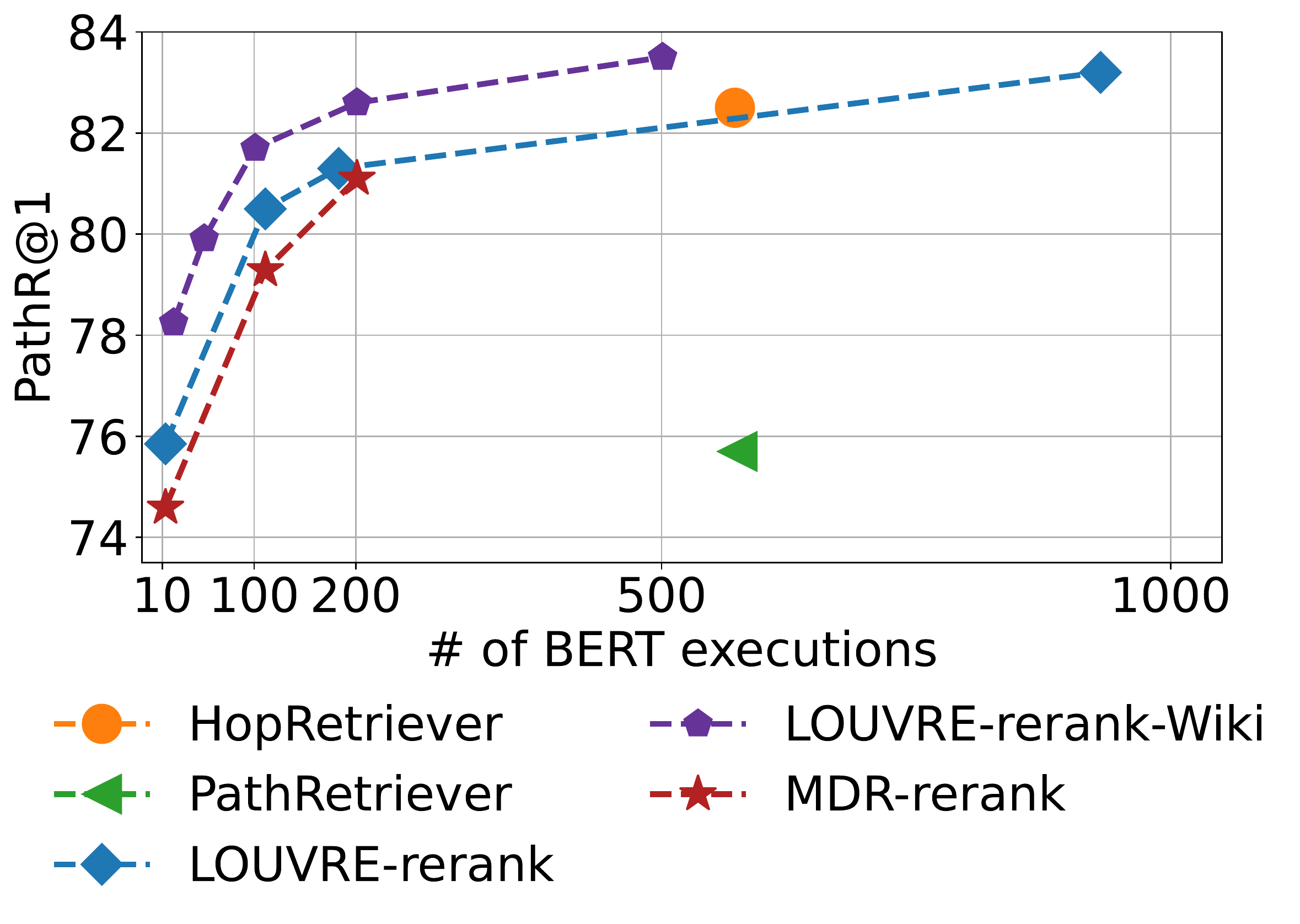}
    \end{subfigure}
    \caption{\attn{
    The retrieval performance of baselines and their computational efficiency. The beam sizes used for each model are as follows: $[3, 10, 20, 40]$ for \modelname{}-eff, $[3, 10, 13, 30]$ for \modelname{}-rerank, $[10, 10, 50, 100, 100]$ for \modelname{}-rerank-wiki, and $[3, 10, 100]$ for MDR-rerank. We use the input size of the reranker of each model as follows: $[9, 100, 169, 900]$ for \modelname{}-rerank, $[10, 40, 50, 100, 400]$ for \modelname{}-rerank-wiki, and $[9, 100, 100]$ for MDR.
    We evaluate PathRetriever with different number of initial n-documents retrieved by TF-IDF; $n \in \{20, 50, 100, 500\}$.
    }}
    \label{fig:time_eff}
\end{figure}

\begin{table*}[t]
\centering
\begin{tabular}{@{}lcccccc@{}}
\toprule
                     & \multicolumn{2}{c}{Answer}      & \multicolumn{2}{c}{Support}     & \multicolumn{2}{c}{Joint}       \\
                     & EM             & F1             & EM             & F1             & EM             & F1             \\ \midrule
SemanticRetrievalMRS & 45.32          & 57.34          & 38.67          & 70.83          & 25.14          & 47.60          \\
Transformer-XH       & 51.60          & 64.07          & 40.91          & 71.42          & 26.14          & 51.29          \\
PathRetriever        & 60.04          & 72.96          & 49.08          & 76.41          & 35.35          & 61.18          \\
MDR           & 62.28          & 75.29          & 57.46          & 80.86          & 41.78          & 66.55          \\ \midrule
\modelname{}        & \textbf{62.90} & \textbf{75.82} & \textbf{57.71} & \textbf{81.26} & \textbf{42.18} & \textbf{67.08} \\ \bottomrule
\end{tabular}
\caption{End-to-end multi-hop QA performance of models on \normalsize{H}\small{OTPOT}\normalsize{QA} test set.}
\label{tab:test}
\end{table*}

\begin{table}[]
\centering
\begin{tabular}{@{}lcc@{}}
\toprule
Retriever           & \#BERT               & Joint F1             \\ \midrule
Reader: BERT        & \multicolumn{1}{l}{} & \multicolumn{1}{l}{} \\
- PathRetriever-eff & $50^+$               & 56.85                \\
- \modelname{}-eff        & $50^+$               & \textbf{60.23}       \\ \midrule
Reader: ELECTRA     & \multicolumn{1}{l}{} & \multicolumn{1}{l}{} \\
- MDR               & $450^+$              & 66.55                \\
- \modelname{}-Wiki       & $450^+$              & \textbf{66.87}       \\ \bottomrule
\end{tabular}
\caption{The end-to-end multi-hop performance of \modelname{}-eff and \modelname{}-Wiki on \normalsize{H}\small{OTPOT}\normalsize{QA} test set with the same inference speed as baselines.}
\label{tab:end2end_w_the_same_inftime}
\end{table}

\attn{Furthermore, we evaluate the zero-shot performance of \modelname{}-eff and compare to DPR not fine-tuned on \normalsize{H}\small{OTPOT}\normalsize{QA}.
To adapt DPR to the multi-hop retrieval task, we replace encoders in \modelname{}-eff with DPR encoders.
We report this result in Table \ref{tab:retriever_eff}.
In Table \ref{tab:retriever_eff}, \modelname{}-zeroshot achieves higher performance (R@10: 44.8 and R@20: 55.8) than DPR-zeroshot (R@10: 39.0 and R@20: 51.6).
}

\noindent\emph{\textbf{Computational Efficiency:}}
We compare the inference time of baselines and \modelname{}, with the number of BERT-base executions needed for each question.
We exclude the inference time for document indexing which can be done a priori.
\attn{The number of BERT executions for each baseline is derived from each paper and its implementation.
We measure the inference time of \modelname{}, PathRetriever, and MDR in various hyper-parameter settings by adjusting the number of beam size and the number of documents retrieved by the sparse retriever, TF-IDF.
The number of BERT executions of MDR, \modelname{}-eff, and \modelname{}-eff with a beam size of $b$ is calculated as follows: $\text{\#BERT} = 1\text{(question encoding)} + b\text{(question-passage encoding)}$.
The inference time of MDR-rerank, \modelname{}-rerank, and \modelname{}-rerank-wiki involves another factor, the input size of the reranker.
The inference time of these reranking models with a beam size of $b$ and a input size of $r$ becomes $\text{\#BERT}=1 + b + r$.
For PathRetriever, we vary the number of documents retrieved by the sparse retriever, TF-IDF.
Figure~\ref{fig:time_eff} illustrates that \modelname{} is more effective and efficient than the baselines because it yields better retrieval performance with a much smaller number of BERT executions.
}

\subsection{End-to-End Multi-Hop QA}
\attn{In this section, we demonstrate that the end-to-end multi-hop QA pipeline using \modelname{} retains the three outcomes of \modelname{}: overall performance improvement, robustness on a few-shot setting, and the fast inference speed.
We use multi-hop QA pipelines of MDR and PathRetriever.
All the components of the multi-hop QA pipelines except the retriever are fixed.
We plug in each baseline retriever and \modelname{} to the pipeline and evaluate the end-to-end performance of each model.
}

\noindent\emph{\attn{\textbf{End-to-End Performance:}}}
\attn{Table \ref{tab:test} shows the end-to-end multi-hop QA performance of baselines and \modelname{}.
In this experiment, we replace the retriever of MDR's pipeline with \modelname{}-rerank.
We set the beam size to 30 and the input size of the reranker to 900.
\modelname{} outperforms baselines with a Joint F1 of 67.08.
Table \ref{tab:end2end_w_the_same_inftime} shows the performance of \modelname{}-Wiki using the same inference time as MDR.
In this experiment, we use a beam size of 100 and the reranker's input size of 350.
MDR uses a beam size of 200 and the reranker's input size of 250.
\modelname{}-Wiki outperforms MDR by 0.32\% point in terms of Joint F1.
We conduct the same experiment with \modelname{}-eff and PathRetriever.
Table \ref{tab:end2end_w_the_same_inftime} shows the results.
\modelname{}-eff in Table \ref{tab:end2end_w_the_same_inftime} represents the end-to-end pipeline of PathRetriever-eff using \modelname{}-eff as the initial candidate document retriever not TF-IDF.
We set the number of initial candidate documents of \modelname{}-eff and PathRetriever-eff to 50.
\modelname{}-eff outperforms PathRetriever-eff by 3.38\% point without any loss of computational efficiency.
We provide the detailed experimental results of \modelname{}-eff in Appendix \ref{sec:appendix_end2end}.
}

\begin{table}[]
\centering
\begin{tabular}{@{}lcc|c@{}}
\toprule
                   & RR         & \#BERT    & Joint F1 \\ \midrule
PathRetriever      & \checkmark & $500^{+}$ & 59.5     \\
- fast inference   & \checkmark & $100^{+}$ & 58.6     \\
- fast inference   & \checkmark & $50^{+}$  & 56.7     \\ \midrule
\modelname{}-eff   & \checkmark & $100^{+}$ & 60.5     \\
- fast inference   & \checkmark & $50^{+}$  & 60.1     \\ \midrule
\modelname{}-eff   &            & 41        & 57.9     \\
- 1\% train set    &            & 41        & 57.1     \\
- w/o pre-training &            & 41        & 54.3     \\
- fast inference   &            & 5         & 53.3     \\ \bottomrule
\end{tabular}
\caption{End-to-end multi-hop QA performances of baseline models and \modelname{} on {\fontsize{10}{12}\selectfont H}{\fontsize{8}{10}\selectfont OTPOT}{\fontsize{10}{12}\selectfont QA} dev set.
\#BERT denotes the number of BERT-base executions in supporting document prediction and the number of documents that the reader takes as input. We exclude the inference time of the supporting sentence selector.
We use the same reader model as PathRetriever.}
\label{tab:abl}
\end{table}

\noindent\emph{\attn{\textbf{End-to-End Performance of Variations of \modelname{}-eff:}}}
\attn{Table \ref{tab:abl} shows the end-to-end performance of each model on different inference time, size of train data, and pre-training.
We evaluate two types of \modelname{}-eff.
\modelname{}-eff (RR) represents the same pipeline used in Table \ref{tab:end2end_w_the_same_inftime}.
\modelname{}-eff (w/o RR) represents the pipeline that consists of \modelname{}-eff and the reader.
Comparison between PathRetriever($500^+$) and \modelname{}-eff ($100^+$) shows that applying \modelname{}-eff to PathRetriever increases Joint F1 by 1 with 5 times faster inference speed.
We conduct the same experiment by reducing the number of documents retrieved by \modelname{}-eff and achieve 0.6\% point higher Joint F1 than PathRetriever with 10 times faster inference speed.
}

\attn{We conduct ablation studies with two factors of retrievers: 1) size of train data and 2) pre-training.
We fix the reader with BERT-wwm fully fine-tuned on \normalsize{H}\small{OTPOT}\normalsize{QA} train set.
Table \ref{tab:abl} shows the results.
When we train \modelname{}-eff with only 1\% of the train set, the end-to-end performance drops by 0.8\% point.
However, using RoBERTa with the whole train set decreases the performance by 3.6\% point.
This result indicates that our pre-training methods bring more significant improvement to the end-to-end multi-hop QA pipeline when the size of the train data is small.
}

\attn{Decreasing the search space of multi-hop retrievers increases the retriever's computational efficiency but results in a significant performance drop.
We demonstrate the robustness of \modelname{}-eff when the computation time is limited.
We decrease the beam size of \modelname{}-eff to 2 and the number of output paths to 2; the total number of BERT executions of this model is $\text{\#BERT} = 1 + 2(\text{beam size}) + 2(\text{input size})$.
Table \ref{tab:abl} shows that \modelname{}-eff achieves 89\% of PathRetriever's performance with almost 100 times faster inference speed of PathRetriever.
We adjust the inference speed of PathRetriever and compare it with \modelname{}-eff (w/o RR).
\modelname{}-eff (w/o RR) outperforms PathRetriever ($50^+$) by 1.2\% point with less computation time.
\modelname{}-eff (w/o RR) trained only with 1\% train data even outperforms PathRetriever ($50^+$).
}
\section{Conclusion}
Answering complex questions includes reasoning across multiple documents.
\attn{Recent studies have found that reasoning requires learning sub-question detection and relevant document retrieval to predict n correct answer with supporting facts.
However, building such datasets requires costly human annotation and has limited scalability.
To address this issue, we proposed a weakly supervised pre-training method for multi-hop retriever, \modelname.
Our pre-training method contains three elements: ``Next Document Prediction" task, ``Bridge Entity Re-Phrasing", and a model.
We demonstrated the efficacy of \modelname{} and its robustness on few-shot settings with extensive experiments on supporting document retrieval task and end-to-end multi-hop QA task.}
We also showed that our method performs very well at a much lower inference cost.
\section*{Acknowledgements}
This work was partly supported by NAVER Corp. and Institute for Information \& communications Technology Planning \& Evaluation(IITP) grant funded by the Korean government(MSIT) (No. 2017-0-01780, The technology development for event recognition/relational reasoning and learning knowledge based system for video understanding).

\bibliographystyle{acl_natbib}
\bibliography{acl2021}

\appendix
\section{Appendix: Baselines}\label{sec:appendix_baselines}

\begin{table*}[t]
\centering
\begin{tabular}{@{}lcccccccc@{}}
\toprule
                   &            &                        & \multicolumn{2}{c}{Answer} & \multicolumn{2}{c}{Support} & \multicolumn{2}{c}{Joint} \\
                   & RR         & \#BERT                 & EM           & F1          & EM           & F1           & EM          & F1          \\ \midrule
PathRetriever      & \checkmark & 500\textsuperscript{+} & 60.2         & 72.8        & 46.3         & 73.7         & 33.8        & 59.5        \\
                   & \checkmark & 100\textsuperscript{+} & 59.7         & 72.6        & 44.8         & 72.6         & 32.7        & 58.6        \\
                   & \checkmark & 50\textsuperscript{+}  & 58.3         & 71.1        & 42.7         & 70.8         & 31.0        & 56.7        \\ \midrule
\modelname{}-eff   & \checkmark & 100\textsuperscript{+} & 60.4         & 73.4        & 48.0         & 74.7         & 35.1        & 60.5        \\
- fast inference   & \checkmark & 50\textsuperscript{+}  & 60.2         & 72.9        & 47.6         & 74.5         & 34.8        & 60.1        \\ \midrule
\modelname{}-eff   &            & 41                     & 59.3         & 71.5        & 44.9         & 71.7         & 33.4        & 57.9        \\
- 1\% train data   &            & 41                     & 58.7         & 71.0        & 44.3         & 70.7         & 32.6        & 57.1        \\
- w/o pre-training &            & 41                     & 57.9         & 70.1        & 42.4         & 69.0         & 31.5        & 54.3        \\
- fast inference   &            & 5                      & 54.3         & 66.3        & 41.9         & 67.2         & 30.5        & 53.3        \\ \bottomrule
\end{tabular}
\caption{End-to-end multi-hop QA performances of baseline models and \modelname{} on {\fontsize{10}{12}\selectfont H}{\fontsize{8}{10}\selectfont OTPOT}{\fontsize{10}{12}\selectfont QA} dev set.}
\label{tab:dev_eff}
\end{table*}

\begin{table*}[]
\centering
\begin{tabular}{@{}lccccccc@{}}
\toprule
                                               &        & \multicolumn{2}{c}{Answer}      & \multicolumn{2}{c}{Support}     & \multicolumn{2}{c}{Joint}       \\
                                               & \#BERT & EM             & F1             & EM             & F1             & EM             & F1             \\ \midrule
SemanticRetrievalMRS                           & 39.4/- & 45.32          & 57.34          & 38.67          & 70.83          & 25.14          & 47.60          \\
Transformer-XH                                 & 100/-  & 51.60          & 64.07          & 40.91          & 71.42          & 26.14          & 51.29          \\
PathRetriever                             & 50\textsuperscript{+}   & 58.21          & 70.86          & 42.91          & 71.30          & 30.95          & 56.85          \\ \midrule
\modelname{}-eff & 50\textsuperscript{+}   & \textbf{59.79} & \textbf{72.65} & \textbf{47.95} & \textbf{74.89} & \textbf{34.54} & \textbf{60.23} \\ \bottomrule
\end{tabular}
\caption{Evaluation results on {\fontsize{10}{12}\selectfont H}{\fontsize{8}{10}\selectfont OTPOT}{\fontsize{10}{12}\selectfont QA} test set. We retrieve the test results of PathRetriever (50\textsuperscript{+}) model and our model from the {\fontsize{10}{12}\selectfont H}{\fontsize{8}{10}\selectfont OTPOT}{\fontsize{10}{12}\selectfont QA} leaderboard.}
\label{tab:test_eff}
\end{table*}

\noindent\emph{\textbf{Retrievers:}}
Retrievers encode questions and documents independently, and search documents by comparing the question vector and the document vectors.
\attn{Retrievers are computationally efficient compared to rerankers since the document vectors can be indexed before the questions are given.}
In \textbf{TF-IDF}, to get the list of supporting documents from the retrieved documents, we rearrange the ranked documents $[d_1, d_2, d_3, d_4, ...]$ to $[\{d_1, d_2\}, \{d_3, d_4\}, ...]$.
\attn{
\textbf{DPR} \cite{karpukhin-etal-2020-dense} is a single-hop document retrieval model.
We compare \modelname{} to the performance of DPR fine-tuned on \normalsize{H}\small{OTPOT}\normalsize{QA} reported from \newcite{xiong2020answering}.
}
\attn{
\textbf{Cognitive Graph} \cite{ding2019cognitive} and \textbf{\normalsize{G}\small{OLD}\normalsize{E}\small{N} \normalsize{Retriever}} \cite{qi2019answering} are multi-hop document retrieval models.
We report the performance of Cognitive Graph reported from \newcite{asai2019learning} and the performance of \normalsize{G}\small{OLD}\normalsize{E}\small{N} \normalsize{Retriever} reported from \newcite{qi2019answering}.
We fine-tune \modelname{} with the same training method proposed in \textbf{MDR} \cite{xiong2020answering}; thus, the performance difference between \modelname{} and MDR only results from our pre-training method.
We report the performance of MDR from \newcite{xiong2020answering}.
}

\noindent\emph{\textbf{Rerankers:}}
\attn{
Rerankers reorder reasoning paths/documents predicted by retrievers.
Rerankers calculate score of each paths/documents by jointly encode each document with the given question.
}
\attn{As a result, reranking takes a huge portion of computation time of the end-to-end multi-hop QA pipeline.}
\attn{
\textbf{SemanticRetrievalMRS} \cite{nie2019revealing} propose the document reranking model that takes output of sparse retrievers such as TF-IDF.
}
Since the model outputs documents not a list of supporting documents, we use the same document rearranging method as TF-IDF above.
\attn{\textbf{PathRetriever} \cite{asai2019learning} and \textbf{HopRetriever} \cite{li2020hopretriever} are reasoning path prediction models.
These models use TF-IDF and BERT to retrieve and rerank the candidate documents.
}
They use Wikipedia hyperlinks for candidate documents selection as described in section \ref{sec:retrieving_strategy} and beam search with size 8 to rank each predicted supporting documents.
\attn{
\textbf{MDR} \cite{xiong2020answering} provides a reranking model as well as their retriever.
We report the performance of MDR-reranking from \newcite{xiong2020answering}.
}

\section{\attn{Appendix: End-to-End Performance of \modelname{}-eff}}\label{sec:appendix_end2end}
\attn{Table \ref{tab:dev_eff} shows the additional results of Table \ref{tab:abl}.
We evaluate \modelname{}-eff on other evaluation metrics used in the \normalsize{H}\small{OTPOT}\normalsize{QA} benchmark and verify the efficacy of \modelname{}-eff.
We report the detailed results of Table \ref{tab:end2end_w_the_same_inftime} in Table \ref{tab:test_eff}.
The results show the end-to-end performance of \modelname{}-eff and PathRetriever on the \normalsize{H}\small{OTPOT}\normalsize{QA} test set.
}

\end{document}